\documentclass{article} 
\usepackage{iclr2018_conference,times}
\usepackage{hyperref}
\usepackage{url}

\usepackage{times}
\usepackage{algorithm}
\usepackage{graphicx}
\usepackage{subcaption}
\usepackage{algorithmic}
\usepackage[utf8]{inputenc} 
\usepackage[T1]{fontenc}    
\usepackage{booktabs}       
\usepackage{amsfonts}       
\usepackage{nicefrac}       
\usepackage{microtype}      
\usepackage{comment}

\title{Large Batch Training of Convolutional Networks}

\author {Yang You  
\thanks{Work was performed when Y.You and I.Gitman were NVIDIA interns } \\
Computer Science Division\\
University of California at Berkeley \\
\texttt{youyang@cs.berkeley.edu} \\
\And 
Igor Gitman\\
Computer Science Department \\
Carnegie Mellon University \\
\texttt{igitman@andrew.cmu.edu} \\
\And Boris Ginsburg\\
NVIDIA \\
\texttt{bginsburg@nvidia.com} \\
}

\iclrfinalcopy 

\begin{document}

\maketitle

\begin{abstract}
A common way to speed up training of large convolutional networks is to add  computational units. Training is then performed using data-parallel synchronous Stochastic Gradient Descent (SGD) with mini-batch divided between computational units. With an increase in the number of nodes, the batch size grows. But training with large batch size often results in the lower model accuracy. We argue that the current recipe for large batch training (linear learning rate scaling with warm-up) is not general enough and training may diverge. To overcome this optimization difficulties we propose a new training algorithm based on Layer-wise Adaptive Rate Scaling (LARS). Using LARS, we scaled Alexnet up to a batch size of 8K, and Resnet-50 to a batch size of 32K without loss in accuracy.

\end{abstract}

\section{Introduction}
\label{sec:intro}

Training of large Convolutional Neural Networks (CNN) takes a lot of time. The brute-force way to speed up CNN training is to add more computational power (e.g. more GPU nodes) and train network using data-parallel Stochastic Gradient Descent, where each worker receives some chunk of global mini-batch (see e.g.  \cite{krizhevsky2014one} or \cite{goyal2017accurate} ).  The size of a chunk should be large enough to utilize the computational resources of the worker. So scaling up the number of workers results in the increase of batch size. But using large batch may negatively impact the model accuracy, as was observed in \cite{krizhevsky2014one},  \cite{li2014efficient}, \cite{keskar2016large}, \cite{hoffer2017train},.. 

 Increasing the global batch while keeping the same number of epochs means that you have fewer iterations to update weights. The straight-forward way to compensate for a smaller number of iterations is to do larger steps by increasing the learning rate (LR). For example,  \cite{krizhevsky2014one} suggests to linearly scale up LR with batch size. However using a larger LR makes optimization more difficult, and networks may diverge especially during the initial phase. To overcome this difficulty,  \cite{goyal2017accurate} suggested doing a "learning rate warm-up": training starts with a small "safe" LR, which is slowly increased to the target "base" LR.  With a LR warm-up and a linear scaling rule,  \cite{goyal2017accurate} successfully trained Resnet-50 with batch B=8K (see also \cite{Cho2017poweraiddl}).    
Linear scaling of LR with a warm-up is the "state-of-the art" recipe for large batch training.

We tried to apply this linear scaling and warm-up scheme to train  Alexnet on Imagenet (\cite{deng2009imagenet}), but scaling stopped after B=2K since training diverged for large LR-s. For B=4K the accuracy dropped from the baseline 57.6\% ( for B=256) to 53.1\%, and for B=8K the accuracy decreased to 44.8\%. To enable training with a large LR, we replaced Local Response Normalization layers in Alexnet with Batch Normalization (BN). We will refer to this modification of AlexNet as AlexNet-BN throughout the rest of the paper. BN improved both model convergence for large LR as well as accuracy: for B=8K  the accuracy gap was decreased from 14\% to 2.2\%.

To analyze the training stability with large LRs we measured the ratio between the norm of the layer weights and norm of gradients update. We observed that if this ratio is too high, the training may become unstable. On other hand, if the ratio is too small, then weights don’t change fast enough. This ratio varies a lot between different layers, which makes it necessary to use a separate LR for each layer. Thus we propose a novel Layer-wise Adaptive Rate Scaling (LARS) algorithm. There are two notable differences between LARS and other adaptive algorithms such as ADAM (\cite{kingma2014adam}) or RMSProp  (\cite{tieleman2012lecture}): first, LARS uses a separate learning rate for each layer and not for each weight, which leads to better stability. And second, the magnitude of the update is controlled with respect to the weight norm for better control of training speed. With LARS we trained Alexnet-BN and Resnet-50 with B=32K without accuracy loss.

\section{Background}
\label{sec:background}
The training of CNN is done using Stochastic Gradient (SG) based methods. At each step $t$ a mini-batch of $B$ samples $x_i$ is selected from the training set. The gradients of loss function  $\nabla L(x_i, w)$ are computed for this subset, and  networks weights $w$ are updated based on this stochastic gradient:
\begin{equation}
    w_{t+1} = w_t - \lambda \frac{1}{B} {\sum}_{i=1}^{B} \nabla L(x_i,  w_t)
    \label{eq:update1}
\end{equation}
The computation of SG can be done in parallel by $N$ units, where each unit processes a chunk  of  the mini-batch with $\frac{B}{N}$ samples. Increasing the mini-batch permits scaling to more nodes without reducing the workload on each unit. However, it was observed that training with a large batch is difficult. To maintain the network accuracy, it is necessary to carefully adjust training hyper-parameters (learning rate, momentum etc). 

\cite{krizhevsky2014one} suggested the following rules for training with large batches: when you increase the batch $B$ by $k$,  you should also increase LR by $k$  while keeping other hyper-parameters (momentum, weight decay, etc) unchanged. The logic behind {\bf linear LR scaling} is straight-forward: if you increase $B$ by $k$ while keeping the number of epochs unchanged, you will do $k$ fewer steps. So it seems natural to increase the step size by $k$. 
For example, let's take $k=2$. The weight updates for batch size $B$ after 2 iterations would be:
\begin{equation}
  w_{t+2} = w_t - \lambda * \frac{1}{B} 
  \bigl( {\sum}_{i=1}^{B} \nabla L(x_i, w_t) +  {\sum}_{j=1}^{B} \nabla L(x_j,  w_{t+1} \bigr)
\end{equation}
The weight update for the batch $B_2=2*B$ with learning rate $\lambda_2$:
\begin{equation}
  w_{t+1} = w_{t} - \lambda_2 * \frac{1}{2*B}  {\sum}_{i=1}^{2B} \nabla L(x_i, w_t)
\end{equation}
will be similar if you take $\lambda_2 = 2* \lambda$, assuming that $\nabla  L(x_j,  w_{t+1}) \approx  L(x_j,  w_t $) . 

Using the "linear LR scaling"  \cite{krizhevsky2014one} trained AlexNet with batch B=1K with minor ($\approx 1\%$) accuracy loss. The scaling of Alexnet above 2K is difficult, since the training diverges for larger LRs. 
It was observed that linear scaling works much better for networks with Batch Normalization (e.g. \cite{codreanu2017xeonphi}). For example   \cite{chen2016revisiting} trained the Inception model with batch B=6400, and  \cite{li2017scaling} trained Resnet-152 for B=5K. 
 
The main obstacle for scaling up batch  is the instability of training with high LR.   \cite{hoffer2017train} tried to use less aggressive "square root scaling" of LR with special form of Batch Normalization ("Ghost Batch Normalization") to train Alexnet with B=8K, but still the accuracy (53.93\%) was much worse than baseline 58\%. 
To overcome the instability during initial phase, \cite{goyal2017accurate}  proposed  to use {\bf LR warm-up}: training starts with small LR, and then LR is gradually increased to the target. After the warm-up period (usually a few epochs), you switch to the regular LR policy ("multi-steps", polynomial decay etc). Using LR warm-up and linear scaling  \cite{goyal2017accurate} trained Resnet-50 with batch B=8K without loss in accuracy. These recipes constitute the current state-of-the-art for large batch training, and we used them as the starting point of our experiments

Another problem related to large batch training is so called "generalization gap", observed by \cite{keskar2016large}. They came to conclusion that "the lack of generalization ability is due to the fact that large-batch methods tend to converge to sharp minimizers of the training function." They tried a few methods to improve the generalization with data augmentation and warm-starting with small batch, but they did not find a working solution.

\section{Analysis of Alexnet training with large batch}
\label{sec:Alexnet}

We used  BVLC\footnote{https://github.com/BVLC/caffe/tree/master/models/bvlc\_alexnet} Alexnet with batch B=512 as baseline. Model was trained using SGD with momentum 0.9 with initial LR=0.01 and the polynomial (power=2) decay LR policy for 100 epochs. The baseline accuracy is 58\% (averaged over last 5 epochs). 
Next we tried to train Alexnet with B=4K by using larger LR. In our experiments we changed the base LR from 0.01 to 0.08, but training diverged with LR > 0.06  even with warm-up \footnote{LR  starts from 0.001 and is linearly increased it to the target LR during 2.5 epochs}. The best accuracy for B=4K is 53.1\%, achieved for LR=0.05. For B=8K we couldn't scale-up LR either, and the best accuracy is 44.8\% , achieved for LR=0.03 (see Table \ref{tab:alexnet}(a) ).

To stabilize the initial training phase we replaced Local Response Normalization layers with Batch Normalization (BN). We will refer to this  model as Alexnet-BN 
\footnote{ https://github.com/borisgin/nvcaffe-0.16/tree/caffe-0.16/models/alexnet\_bn}. The baseline accuracy for Alexnet-BN with B=512 is 60.2\%. 
\footnote{
Alexnet-BN baseline was trained using SGD with  momentum=0.9, weight decay=0.0005 for 128 epochs. We used polynomial (power 2) decay LR policy with base LR=0.02. }
With BN we could use large LR-s even without warm-up.  For B=4K the best accuracy 58.9\% was achieved for LR=0.18,  and for B=8K the best accuracy 58\%  was achieved  for LR=0.3. We also observed that BN significantly widen the range of LRs with good accuracy.

\begin{figure*}[htb!]
\centering
\includegraphics[width=0.7\textwidth]{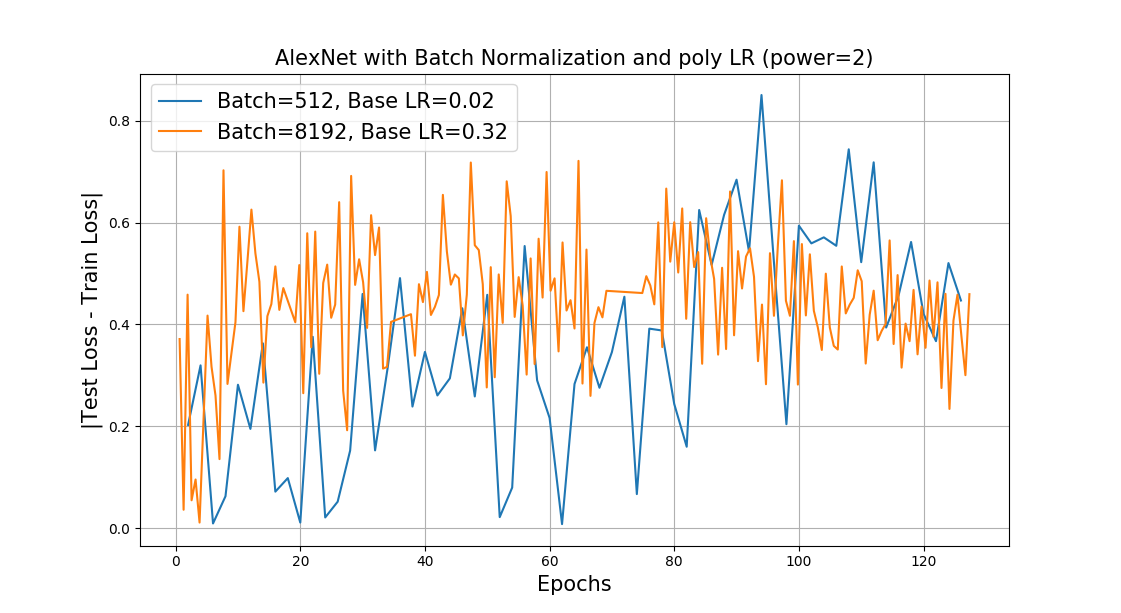}
\caption{Alexnet-BN: Gap between training and testing loss}
\label{fig:alexnetbn_loss_diff}
\vspace{-10pt}
\end{figure*}

\begin{table}[htb!]
  \caption{Alexnet and Alexnet-BN: B=4K and 8K. BN makes it possible to use larger learning rates.}
  \label{tab:alexnet}
  \centering
  {\renewcommand{\arraystretch}{1.2}
  \begin{tabular}[t]{|c|c|c|}
  \multicolumn{3}{c}{(a) Alexnet (warm-up 2.5 epochs)} \\
  \hline 
  Batch & Base LR & accuracy,\% \\
  \hline
  {\bf 512 } & {\bf 0.02}  &  {\bf 58.8} \\
  \hline
  4096 & 0.04 &  53.0 \\
  {\bf 4096} &  {\bf 0.05} & {\bf 53.1} \\
  4096 & 0.06 & 51.6\\
  4096 & 0.07 & 0.1\\
  \hline
  8192 & 0.02 & 29.8 \\
  {\bf 8192} & {\bf 0.03} & {\bf 44.8}\\
  8192 & 0.04 & 43.1 \\
  8192 & 0.05 & 0.1  \\
  \hline
  \end{tabular}}
  \quad
  {\renewcommand{\arraystretch}{1.2}
  \begin{tabular}[t]{|c|c|c|}
  \multicolumn{3}{c}{(b) Alexnet-BN (no warm-up)} \\
  \hline 
  Batch & Base LR  & accuracy,\% \\
  \hline
  {\bf 512} &  {\bf 0.02} & {\bf 60.2}\\
  \hline
  4096 & 0.16 & 58.1 \\
  {\bf 4096} & {\bf 0.18}  & {\bf 58.9}\\
  4096 & 0.21  & 58.5\\
  4096 & 0.30  & 57.1\\
  \hline
  8192 & 0.23 & 57.6 \\
  {\bf 8192} &  {\bf 0.30} & {\bf 58.0} \\
  8192 & 0.32 &  57.7 \\
  8192 & 0.41 &  56.5\\
  \hline 
  \end{tabular}}
\end{table}
Still there is  a $2.2\%$ accuracy loss for B=8K. To check if it is related to the "generalization gap" (\cite{keskar2016large}), we looked at the loss gap between training and testing (see Fig. \ref{fig:alexnetbn_loss_diff}). We did not find the significant difference in the loss gap between B=256 and B=8K. We conclude that in this case the accuracy loss is not related to a generalization gap, and it is caused by the low training.

\section{Layer-wise Adaptive Rate Scaling (LARS)}

The  standard SGD uses the same LR $\lambda$ for all layers:  $w_{t+1} = w_t - \lambda \nabla L(w_t)$. When $\lambda$ is large, the update  $||\lambda * \nabla L(w_t)||$ can become larger than  $||w||$, and this can cause the divergence. This makes the initial phase of training highly sensitive to the weight initialization and to initial LR. We found that the ratio the L2-norm of weights and gradients $||w|| /|| \nabla  L(w_t)||$ varies significantly between weights and biases, and between different layers. For example, let's take AlexNet-BN after one iteration (Table \ref{tab:weight_gradient}, "*.w" means layer weights, and "*.b" - biases). The ratio $||w|| / ||\nabla L(w)||$ for the 1st convolutional layer ("conv1.w") is 5.76, and for the last fully connected layer ("fc6.w") - 1345.
 \begin{table}[htb!]
 \caption{AlexNet-BN: The norm of weights and gradients at 1st iteration.
}
 \label{tab:weight_gradient}
 \centering
 \vspace{3pt}
 \begin{tabular}{|c|cc |cc|cc|cc|}
 \hline Layer  & conv1.b & conv1.w & conv2.b & conv2.w & conv3.b & conv3.w & conv4.b & conv4.w \\
  \hline $||w||$
    & 1.86 & 0.098 & 5.546 & 0.16  & 9.40  & 0.196  & 8.15  & 0.196  \\
  \hline $||\nabla L(w)||$
    & 0.22 & 0.017 & 0.165 & 0.002 & 0.135 & 0.0015 & 0.109 & 0.0013 \\
  \hline  $\frac{||w||} { ||\nabla L(w)||}$  
    & 8.48 & {\bf 5.76} & 33.6 & 83.5 & 69.9 & 127 & 74.6 & 148 \\
  \hline
  \hline Layer & conv5.b & conv5.w & fc6.b & fc6.w & fc7.b & fc7.w & fc8.b & fc8.w\\
  \hline $||w||$ 
    & 6.65 & 0.16 & 30.7 & 6.4 & 20.5 & 6.4 & 20.2 & 0.316\\
  \hline  $||\nabla L(w)||$ 
    & 0.09 & 0.0002 & 0.26 & 0.005 & 0.30 & 0.013 & 0.22 & 0.016\\
   \hline  $\frac{||w||} { ||\nabla L(w)||}$ 
    & 73.6 & 69 & 117 & {\bf 1345} & 68 & 489 & 93 & 19\\
  \hline
  \end{tabular}
\end{table}
The ratio is high during the initial phase, and it is rapidly decrease after few epochs (see Figure 2). If LR is large comparing to the ratio for some layer, then training may becomes unstable. The LR "warm-up" attempts to overcome this difficulty by starting from small LR, which can be safely used for all layers, and then slowly increasing it until weights will grow up enough to use larger LRs.
 
We would like to use different approach. We use local LR  $\lambda^l$ for each layer $l$:
\begin{equation}
     \triangle w^l_t  =  \gamma* \lambda^l * \nabla L(w^l_t) 
\end{equation}
where $\gamma$ is a global LR.  Local LR  $\lambda^l$ is defined for each layer through "trust" coefficient $\eta < 1$:
\begin{equation}
    \lambda^l =  \eta  \times \frac{||w^l||}{||\nabla L(w^l)||}
    \label{eq:lars}
\end{equation}
The $\eta$  defines  how much we trust  the layer to change its weights during one update
\footnote{
One can consider LARS as a private case of block-diagonal re-scaling from \cite{lafond2017diagonal}. 
}.  
Note that now the magnitude of the update for each layer doesn't depend on the magnitude of the gradient anymore, so it helps to partially eliminate vanishing and exploding gradient problems. This definition  can be easily extended for SGD to balance the local learning
rate and the weight decay term  $\beta$:
\begin{equation}
    \lambda^l = \eta  \times \frac{||w^l||} {||\nabla L(w^l)|| + \beta *||w^l|| }
    \label{eq:lars_wd}
\end{equation}

\begin{algorithm}[htb!]
\begin{algorithmic}
\STATE {\bf Parameters:} base LR $\gamma_0$, momentum $m$, weight decay $\beta$, LARS coefficient $\eta$, number of steps $T$
\STATE {\bf Init:} $t = 0, v = 0$. Init weight $w_0^l$ for each layer $l$
\WHILE {$t < T$ for each layer $l$} 
        \STATE $g_t^l \gets \nabla L(w_t^l)$   (obtain a stochastic gradient for the current mini-batch)
        \STATE $\gamma_t \gets \gamma_0 * \left(1 - \frac{t}{T}\right)^2$ (compute the global learning rate)
        \STATE $\lambda^l \gets \frac{||w_t^l||}{||g_t^l|| + \beta ||w_t^l||}$       (compute the local LR  $\lambda^l$)
        \STATE $v_{t+1}^l \gets mv_t^l + \gamma_{t+1} * \lambda^l * (g_t^l + \beta w_t^l)$     (update the momentum)
        \STATE $w_{t+1}^l \gets w_t^l - v_{t+1}^l$ (update the 
        weights)
\ENDWHILE
\end{algorithmic}
 \caption{SGD with LARS. Example with weight decay, momentum and polynomial LR decay. \label{algo:lars}}
\end{algorithm}
The network training for SGD with LARS are summarized in the Algorithm \ref{algo:lars}. One can find more implementation details  at \href{https://github.com/borisgin/nvcaffe-0.16}{https://github.com/borisgin/nvcaffe-0.16}

\begin{figure*}[htb!]
    \begin{subfigure}[b]{0.5\textwidth}
        \centering
        \includegraphics[height=2.1in]{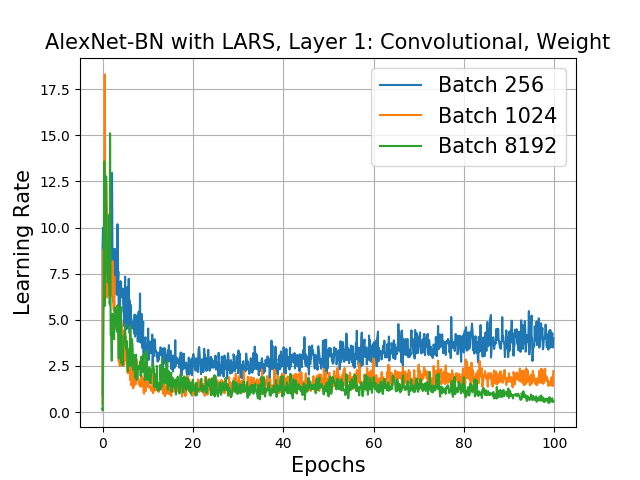}
        \caption{Local LR, conv1-weights}
    \end{subfigure}%
    \begin{subfigure}[b]{0.5\textwidth}
        \centering
        \includegraphics[height=2.1in]{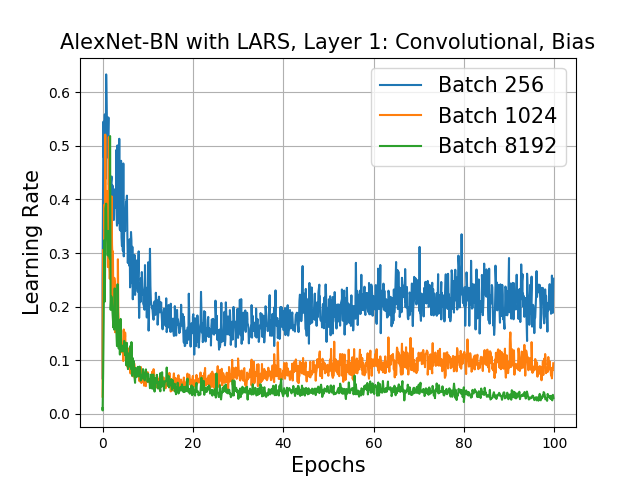}
        \caption{Local LR, conv1-bias}
    \end{subfigure}
    \begin{subfigure}[b]{0.5\textwidth}
        \centering
        \includegraphics[height=2.1in]{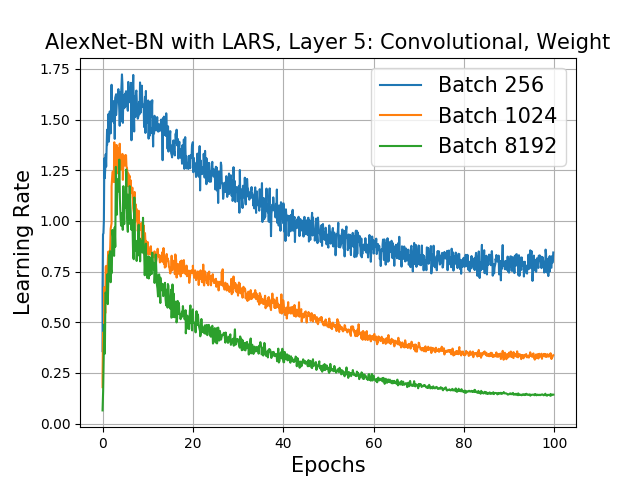}
        \caption{Local LR , conv5-weights}
    \end{subfigure}
    \begin{subfigure}[b]{0.5\textwidth}
        \centering
        \includegraphics[height=2.1in]{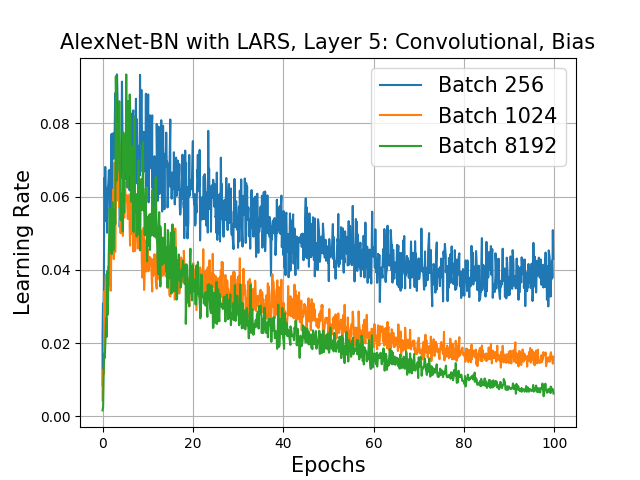}
        \caption{Local LR, conv5-bias}
    \end{subfigure}
    \caption{\label{fig:lars_localLR} LARS: local LR for different layers and batch sizes}
\end{figure*}

The local LR strongly depends on the layer and batch size (see Figure. \ref{fig:lars_localLR} )

\section{Training with LARS}
\begin{figure*}[htb!]
    \begin{subfigure}[b]{0.55\textwidth}
        \centering
        \includegraphics[height=2.5in]{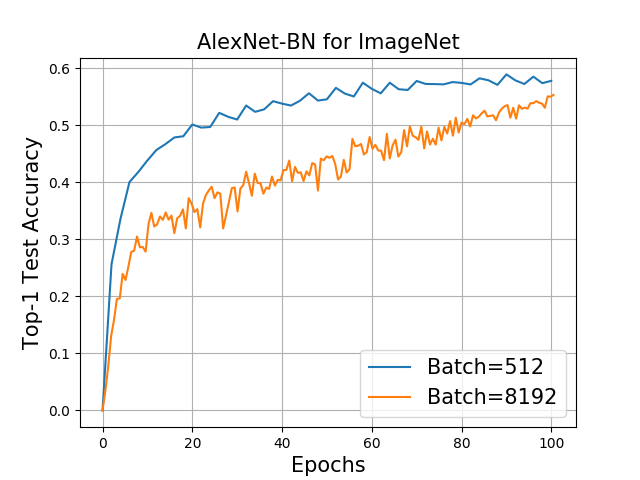}
        \caption{Training without LARS}
    \end{subfigure}
    \begin{subfigure}[b]{0.55\textwidth}
        \centering
        \includegraphics[height=2.5in]{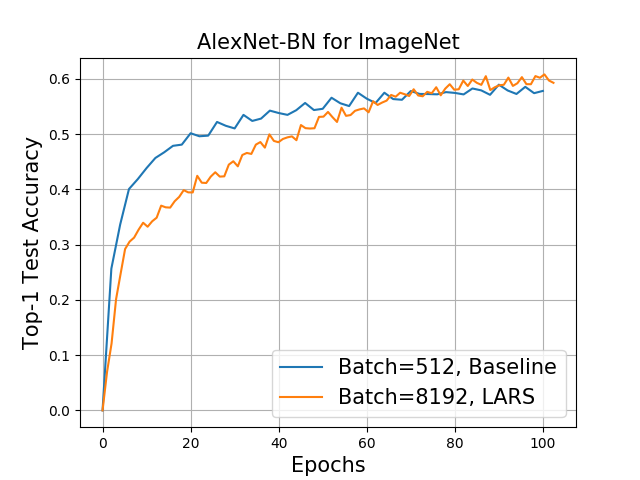}
        \caption{Training with LARS}
    \end{subfigure}
    \caption{\label{fig:alexnetbn_b8k_lars} LARS: Alexnet-BN with B=8K}
\end{figure*}

We re-trained Alexnet and Alexnet-BN with LARS for batches up to 32K 
\footnote{ Models have been trained for 100 epochs using SGD with momentum=0.9, weight decay=0.0005, polynomial (p=2) decay LR policy, and LARS coefficient $\eta=0.001$. Training have been done on NVIDIA DGX1. 
To emulate large batches (B=16K and 32K) we used $iter\_size$ parameter to partition mini-batch into smaller chunks. The weights update is done after gradients for the last chunk are computed.
}. 
For B=8K  the accuracy of both networks matched the baseline B=512 (see Figure \ref{fig:alexnetbn_b8k_lars}). Alexnet-BN trained with B=16K lost 0.9\% in accuracy, and trained with B=32K lost ~2.6\%.
\begin{table}[htb!]
  \caption{Alexnet and Alexnet-BN: Training with LARS}
  \label{tab:alexnet_lars}
  \centering
  {\renewcommand{\arraystretch}{1.2}
  \begin{tabular}[t]{|c|c|c|}
  \multicolumn{3}{c}{(a) Alexnet (warm-up for 2 epochs)} \\[5px]
  \hline Batch & LR  & accuracy,\% \\
  \hline  512  & 2   &  58.7 \\
  \hline  4K   & 10  &  58.5 \\
  \hline  8K   & 10  &  58.2 \\
  \hline 16K   & 14  &  55.0 \\
  \hline 32K   & TBD &  TBD \\
  \hline
  \end{tabular}}
  \quad
  {\renewcommand{\arraystretch}{1.2}
  \begin{tabular}[t]{|c|c|c|}
  \multicolumn{3}{c}{(b) Alexnet-BN (warm-up for 5 epochs)} \\[5px]
  \hline Batch & LR & accuracy,\% \\
  \hline  512  & 2  & 60.2 \\
  \hline  4K   & 10 & 60.4 \\
  \hline  8K   & 14 & 60.1 \\
  \hline 16K   & 23 & 59.3 \\
  \hline 32K   & 22 & 57.8 \\
  \hline
  \end{tabular}}
\end{table}

There is a relatively wide interval of base LRs which  gives the "best" accuracy. for example, for Alexnet-BN with B=16K LRs from [13;22] give the  accuracy $\approx 59.3$, for B=32k, LRs from [17,28] give  $\approx 57.5$
\begin{figure*}[htb!]
\centering
\includegraphics[width=0.7\textwidth]{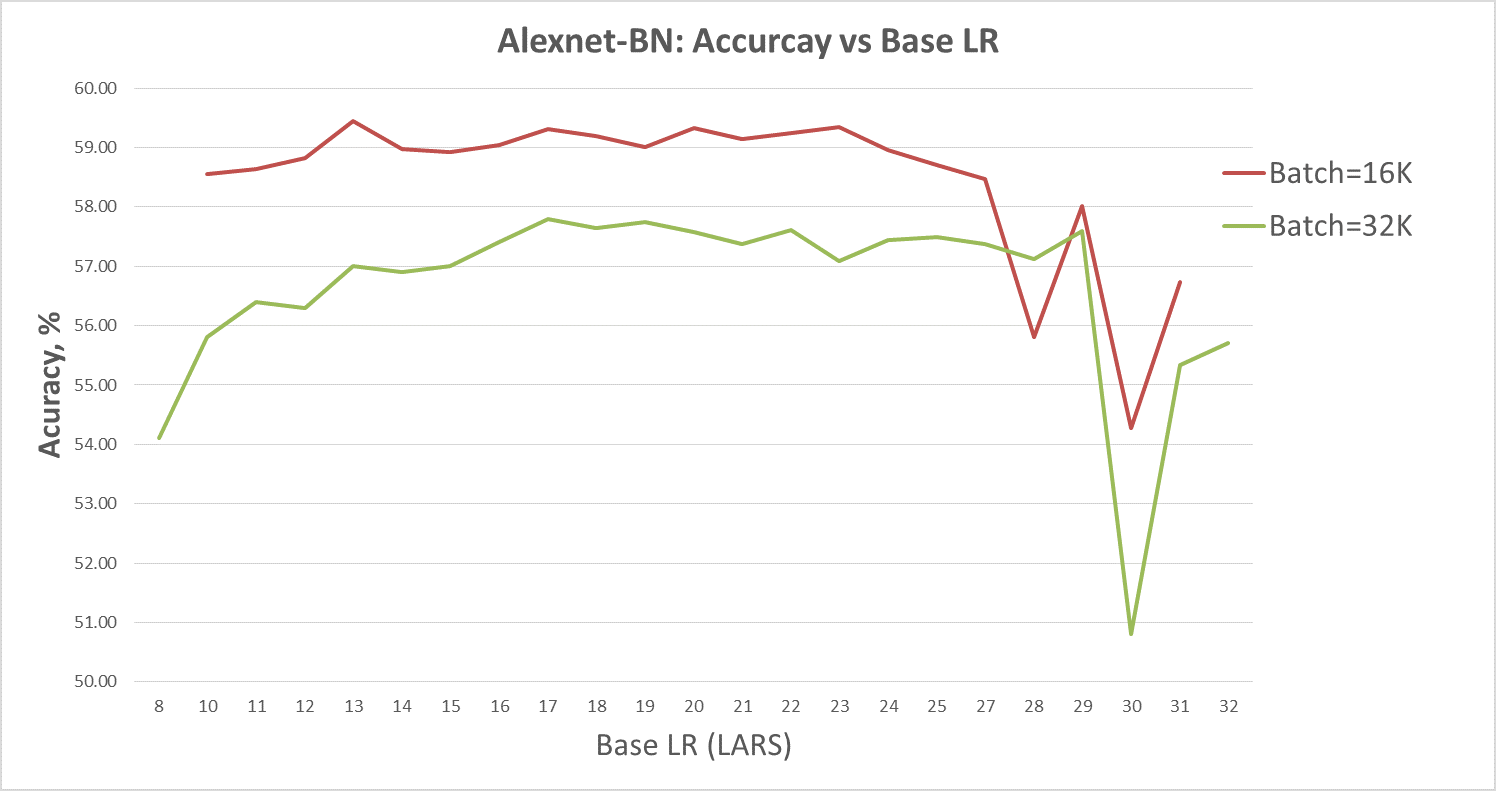}
\caption{Alexnet-BN, B=16K and 32k: Accuracy as function of LR}
\label{fig:alexnetbn_acc_baselr}
\vspace{-10pt}
\end{figure*}

Next we retrained Resnet-50, ver.1 from \cite{he2016deep} with LARS. As a baseline we used B=256 with corresponding top-1 accuracy 73\%. 
\footnote {
Note that our baseline 73\% is lower than the published state-of-the-art 75\% \cite{goyal2017accurate} and \cite{Cho2017poweraiddl} for few reasons. We trained with the minimal data augmentation (pre-scale images to 256x256 and use random 224x224 crop with horizontal flip). During testing we used one model and 1 central crop. The state-of-the art accuracy 75\% was achieved with more extensive data augmentation during testing, and with multi-model, multi-crop testing. For more details see log files  \href{https://people.eecs.berkeley.edu/~youyang/publications/batch}{https://people.eecs.berkeley.edu/$\sim$youyang/publications/batch}.
}

\begin{table}[htb!]
  \caption{ResNet50 with LARS.}
  \label{tab:resnet50_auto_lr}
  \centering
  \vspace{3pt}
  \begin{tabular}{|c|c|c|c|c|}
    \hline
    Batch      & LR policy    &  $\gamma$ & warm-up & accuracy, \% \\
    \hline 256 &      poly(2) &      0.2     & N/A & 73.0 \\
    \hline
    \hline  8K & LARS+poly(2) &      0.6     & 5 & 72.7 \\
    \hline 16K & LARS+poly(2) &      2.5     & 5 & 73.0 \\
    \hline 32K & LARS+poly(2) &      2.9     & 5 & 72.3 \\
    \hline
  \end{tabular}
\end{table}
All networks have been trained using SGD with momentum 0.9 and weight decay=0.0001 for 90 epochs. We used LARS and warm-up for 5 epochs with polynomial decay (power=2) LR policy.
\begin{figure*}[tb]
 \vspace{5pt}
 \centering
 \includegraphics[width=0.88\textwidth]{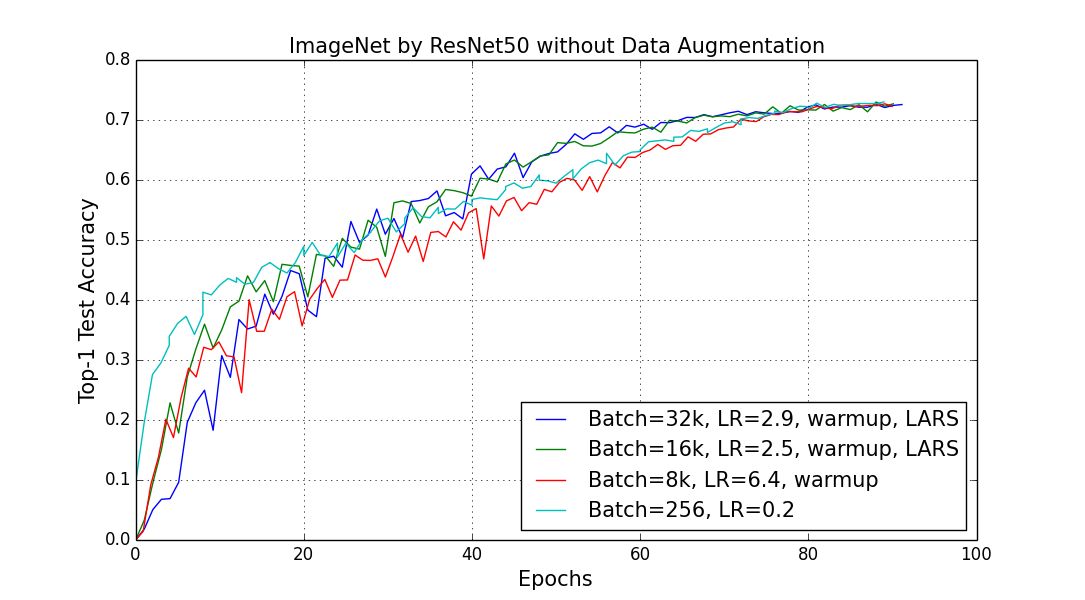}
 \caption{\label{fig:resnet50_32k}Scaling ResNet-50 up to B=32K with LARS.}
 \vspace{-10pt}
\end{figure*}

 We found that with LARS we can scale up Resnet-50 up to batch B=32K with almost the same (-0.7\%) accuracy as baseline
 
\section{Large Batch vs Number of steps}
As one can see from Alexnet-BN exmaple for B=32K, even training with LARS  and using large LR does not reach baseline accuracy. But the accuracy can be recovered completely by just training longer. We argue that when batch very large, the stochastic gradients become very close to true gradients, so increasing the batch does not give much additional gradient information comparing to smaller batches. 

\begin{table}[htb!]
  \caption{Alexnet-BN, B=32K: Accuracy vs Training duration}
  \label{tab:alexnetbn_acc_vs_epochs}
  \centering
  \vspace{3pt}
  \begin{tabular}{|c|c|}
  \hline Num of epochs & accuracy, \% \\
  \hline  100   &  57.8 \\
  \hline  125   &  59.2 \\
  \hline  150   &  59.5 \\
  \hline  175   &  59.5 \\
  \hline  200   &  59.9 \\
  \hline
  \end{tabular}
\end{table}

\section{Conclusion}
\label{sec:conclusion}
Large batch is a key for scaling up training of convolutional networks. 
The existing approach for large-batch training, based on using large learning rates, leads to divergence, especially during the initial phase, even with learning rate warm-up.  To solve these optimization  difficulties we proposed the new algorithm, which adapts the learning rate  for each layer (LARS). Using LARS, we extended scaling of Alexnet and Resnet-50 to B=32K. Training of these networks with batch above 32K without accuracy loss is still open problem.

\bibliographystyle{iclr2018_conference}
\bibliography{iclr2018_conference}

\end{document}